\documentclass{article}

\usepackage{color}
\usepackage{arxiv}

\usepackage[utf8]{inputenc} 
\usepackage[T1]{fontenc}    
\usepackage{hyperref}       
\usepackage{url}            
\usepackage{booktabs}       
\usepackage{amsfonts}       
\usepackage{nicefrac}       
\usepackage{microtype}      
\usepackage{graphicx}
\usepackage{hyperref}
\usepackage{tikz}
\usepackage{pgfplots}
\usetikzlibrary{matrix, positioning}
\usepackage{bm}
\usepackage{relsize}

\usepackage{array, makecell}
\usepackage{xfrac}
\usepackage{multirow}

\usepackage[justification=centering]{caption}
\usepackage{neuralnetwork}

\newcommand{\x}[2]{$x_#2$}
\newcommand{\y}[2]{$\hat{y}_#2$}

\usepackage[ruled]{algorithm2e}

\newcommand{\addimghere}[4]{
    \begin{figure}[H]
        \centering
        \includegraphics[width=#2\linewidth]{#1}
        \caption{#3} \label{#4}
    \end{figure}
}
\newcommand{\addtwoimghere}[6]{
\begin{figure}[H]
    \centering
    \subcaptionbox{#2}{\includegraphics[width=.49\textwidth]{#1}}
    \subcaptionbox{#4}{\includegraphics[width=.49\textwidth]{#3}}
    \caption{#5}
    \label{#6}
\end{figure}
}
\usepackage[labelformat=simple, labelsep=period]{subcaption}
\newcommand{\addthreeimghere}[8]{
    \begin{figure}[H]
        \centering
        \subcaptionbox{#2}{\includegraphics[width=.3\textwidth]{#1}}
        \subcaptionbox{#4}{\includegraphics[width=.3\textwidth]{#3}}
        \subcaptionbox{#6}{\includegraphics[width=.3\textwidth]{#5}}
        \caption{#7}
        \label{#8}
    \end{figure}
}

\usepackage{enumerate}
\usepackage{indentfirst}
\usepackage{float}
\usepackage{multirow}

\usepackage{graphicx}
\graphicspath{{images/}}
\usepackage{etoolbox}

\usepackage{amssymb}
\title{A Comparison of Deep Machine Learning Algorithms in \emph{COVID-19} Disease Diagnosis}

\author{
  Samir S.~Yadav \\
  Department of Information Technology\\
  Dr. Babasaheb Ambedkar Technological University\\
  Lonere, Raigad 402103 \\
  \texttt{ssyadav@dbatu.ac.in} \\
   \And
Jasminder Kaur Sandhu \thanks{Corresponding author:(https://curin.chitkara.edu.in/profile/jasminder-kaur-sandhu/).}   \\
 Chitkara University Institute of Engineering and Technology\\
 Chitkara University\\
 Punjab, India\\
  \texttt{jasminder.sandhu@chitkara.edu.in} \\
  \And
 Mininath R. Bendre  \\
  Department of Computer Engineering\\
  Pravara Rural Engineering College\\
 Loni,Pravaranagar-413736\\
  \texttt{mininath.bendre@gmail.com } \\
   \AND
   Pratap S Vikhe \\
   Department of Instrumentation Engineering\\
  Pravara Rural Engineering College\\
 Loni,Pravaranagar-413736\\
   \texttt{pratapvikhe@gmail.com} \\
   \And
   Amandeep Kaur \\
   Department of Computer Science and Engineering\\
  Sri Guru Granth Sahib World University\\
  Fatehgarh Sahib, Punjab \\
  \texttt{amandeep1426@sggswu.edu.in} \\
}

\begin{document}
\maketitle

\begin{abstract}
The objective of this work is to use deep neural network models for solving the problem of image recognition. These days, every human being is threatened by a harmful coronavirus disease, also called COVID-19 disease. The spread of coronavirus affects the economy of many countries in the world. To find COVID-19 patients early is very essential to avoid the spread and harm to society. Pathological tests and Chromatography(CT) scans are helpful for the diagnosis of COVID-19. However, these tests are having drawbacks such as a large number of false positives, and cost of these tests are so expensive. Hence, it requires finding an easy, accurate, and less expensive way for the detection of the harmful COVID-19 disease. Chest-x-ray(CXR) can be useful for the detection of this disease. Therefore, in this work, CXR images are used for the diagnosis of suspected COVID-19 patients using modern machine learning techniques such as Convolutional Neural network(CNN). CNN models like Inception, ResNet, and DenseNet with three different optimizers(Stochastic Gradient Descent (SGD), RMSProp, and Adam)  are trained and validated to classify Normal, pneumonia, COVID-19 CXR images obtained from online repository. The results of these analyses conclude that the Inception V3 model with the SGD optimizer effectively solve the  CXR image classification problem. 
\end{abstract}


\section{Introduction}
\label{intro}
The new COVID-19 disease caused by the new strain of corona virus was first detected in Wuhan, China since December 2019 \cite{WHOcoronavirus}. The disease subsequently spread throughout China and around the world \cite{WHOoutbreaknews,WHOsituation}. A normal person can become infected if they have close contact with an infected person. Signs and symptoms reported include fever, fatigue, dry cough, shortness of breath, and respiratory failure. Most patients have mild symptoms and have a good prognosis. Cases of death are often the elderly and have underlying medical conditions such as cardiovascular disease and diabetes \cite{WHOcoronavirus}. As of August 15, 2020, 2,589,208 cases and 63896 deaths were reported in India \cite{worldmeter}.\\

The Indian government has used many social isolation and quarantine measures to reduce the spread of disease in the community. Typically search, isolate and track 14 days of people in direct contact with the sick. From 24 March 2020, outbreaks of places like Mumbai, Pune have been completely sealed off. Schools and businesses are suspended to prevent the spread of disease. People in all countries do not get a VISA to enter India. Many domestic and foreign flights were also canceled due to the epidemic.\\
To perform testing for the detection of COVID-19  are challenge for all, particularly the developed countries, because the testing device and its testing kits are quite limited and not accessible worldwide~\cite{sohrabi2020world}. Several investigators and research institutions are currently researching on COVID-19 diagnosis~\cite{yang2020epidemiological,chen2020epidemiological,huang2020clinical,chu2020molecular,yadav2020chest}. Investigating early COVID-19 signs is not a reliable screening technique because there are certain instances in which patients have the signs but not diagnosed with COVID -19 as verified by the clinical examination or CT scan.  While one of the important methods of diagnosing COVID-19 cases is the pathological examination. Nonetheless, this procedure has certain drawbacks because it should be performed in clinical labs that are only located in city centers and need time-consuming results.  This can cause a problem as the positive patients cannot be isolated earlier, and they can infect more people through the crucial time of unrestricted movement ~\cite{wang2020deep,corman2020detection}. CT scan is also one of the popular methods for diagnosis of COVID-19 \cite{ng2020imaging}. However, the problem in this radiological imaging is the overlapping with other diseases. When a COVID-19 patient is infected with another lung disease such as pneumonia, then it is difficult for a medical professional or a radiologist to diagnose these both similar looking CT scan images \cite{kanne2020essentials}. Also, one of the major disadvantages of COVID-19 diagnosis using a CT scan is its high radiation dose and its high cost, as it is not easy for common people to use this procedure ~\cite{fred2004drawbacks}.  Traditional radiography or Chest X-Ray (CXR) images can overcome the problem of costly CT scans and pathological tests because CXR is less costly and has minimal harmful consequences. Also, CXR is capable of identifying various lung diseases earlier ~\cite{bhalla2015chest} and pneumonia diseases~\cite{yadav2019deep}. The CXR imaging is a non-invasive procedure that takes  2-3 minutes to capture the image, and results can be fetched within thirty minutes. The modern radiographical machines are affordable for average income countries or underdeveloped countries ~\cite{ngoya2016defining}. Hence, in this research work, CXR is used to identify the deadly COVID-19 coronavirus.  \\

Currently, due to the rapid development of digital technologies, the use of automated and robotic systems has spread to many areas, both in industry, science and in everyday life. As a consequence, there is an increasing need for efficient processing of information presented, in particular, in video and image formats~\cite{sze2017efficient}. At the current moment, images have closely merged into human life. Therefore, many automated systems use them as the main source of information~\cite{jung2004text}. Finding, localizing, classifying and analyzing objects in an image by a computer is a complex task in computer vision. Computer vision is a set of software and technical solutions in the field of artificial intelligence (AI), aimed at reading and receiving information from images, in real time and without human intervention~\cite{remagnino2004ambient}.
In the process of analysing information received from the eyes, the human brain does a tremendous amount of work.  A person can easily describe what is happening in a randomly taken photograph. Images can carry a tremendous amount of detail and differ in many parameters, such as resolution, color, quality, brightness, noise, etc. Objects in images can also have many features: scale, position, color, rotation, tilt, etc. However, in digital format, each image is just an array of numerical data. Teaching a computer to find and classify images in an image taking into account all factors is a very complex algorithmic task. To solve it, machine learning technologies are actively used~~\cite{yadav2019machine,yadav2019comparative,yadav2020automated,yadav2020application,yadav2020detection}. A person receives a large amount of information through sight.
Images are capable of storing a huge amount of it. As a consequence, their use in computer systems increases the performance of these systems. However, such technologies require complex calculations. The challenge in computer vision is to develop efficient algorithms that extract and analyze data from images or videos.\\
In this work, deep machine learning algorithms are used for solving image recognition problems, and also neural networks are designed and trained for diagnosing COVID-19 from chest x-rays.

Remaining of this paper organized as: section~\ref{ML} gives the basic information about machine learning techniques. Section~\ref{CNN} contains a description of the architecture of a convolutional neural network and a description of modern models based on it. In section~\ref{dev}, modern models of convolutional neural networks were trained to solve the problem of diagnosing pneumonia and COVID-19 using X-ray images. Finally conclusions are drawn in the section~\ref{con}
\section{Technical Background}\label{ML}
\subsection{The concept of an artificial neural network }
Machine learning(ML) is a branch of research in the field of AI, which is based on methods of developing systems capable of learning. Ml algorithms show themselves effectively in tasks in which it is required to determine common features from previously trained data and identify new data from them. Artificial neural networks are often used in the design of such learning systems.

Artificial neural network (ANN) is a computer model, which is based on the principles of a biological neural network - a set of interconnected nerve cells - called as neurons. ~Typical structure of a biological neuron is as shown in figure~\ref{biological-neuron}. Each neuron has a set of input connections - synapses, through which it receives information presented in the form of impulses from other neurons. According to the data obtained, the neuron forms its state and, with the help of the axon, communicates it to other neurons, ensuring the functioning of the system. In the process of forming the system, some neural connections are strengthened, while others are weakened, ensuring the learning of the network.
\addimghere{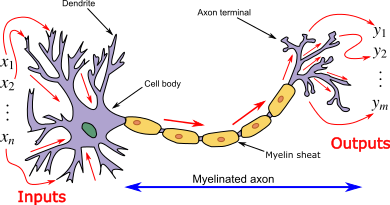}{0.5}{Typical structure of a biological neuron~\cite{wiki:bn}}{biological-neuron}

\begin{figure}[h]
  \centering

  \begin{tikzpicture}
    \tikzstyle{rectangle_style}=[rectangle, draw]
    \tikzstyle{divided rectangle_style}=[draw, rectangle split, rectangle split parts=2, rotate = 90, minimum height = 15mm, minimum width = 10mm]

    \foreach \x in {0,...,2}
      \draw node at (0, -\x) [rectangle_style] (neuron_i_\x) {$x_\x$};
    \foreach \x in {1,...,3}
      \fill (0, -2.5 - \x*0.15) circle (1pt);
    \draw node at (0, -3.5) [rectangle_style] (neuron_i_3) {$x_i$};

    \foreach \x in {0,...,2}
      \draw node at (1.5, -\x) [] (w_ji_\x) {$w_{j\x}$};
    \draw node at (1.5, -3.5) [] (w_ji_i) {$w_{ji}$};
    \foreach \x in {1,...,3}
      \fill (1.5, -2.5 - \x*0.15) circle (1pt);

    \draw node at (4.5, -1.5) [rectangle_style] (neuron_sum) {
      $s_j = \sum {w_{ji}x_i+b_j}$
    };
    \draw node at (7.8, -1.5) [rectangle_style] (neuron_act) {
      $out_j = \varphi (s_j)$
    };
    \foreach \x in {1,...,3}
      \fill (6.4, -2.25 - \x*0.15) circle (1pt);
    \foreach \x in {1,...,3}
      \fill (6.4,  - 0.15 - \x*0.15) circle (1pt);

    \node at (10, -1.5) [circle, draw] (output) {out};

    \foreach \i in {0,...,2}
      \path[-] (neuron_i_\i) edge node[] {} (w_ji_\i);
    \path[-] (neuron_i_3) edge node[] {} (w_ji_i);

    \foreach \i in {0,...,2}
      \path[->] (w_ji_\i) edge node[] {} (neuron_sum);
    \path[->] (w_ji_i) edge node[] {} (neuron_sum);

     \path[->] (neuron_sum) edge node[above, midway] {$ $} (neuron_act);

    \path[->] (neuron_act) edge node[above, midway] {$ $} (output);

  \end{tikzpicture}
\caption{Artificial neuron schematic~\cite{wiki:act}} \label{simple-neuron}
\end{figure}
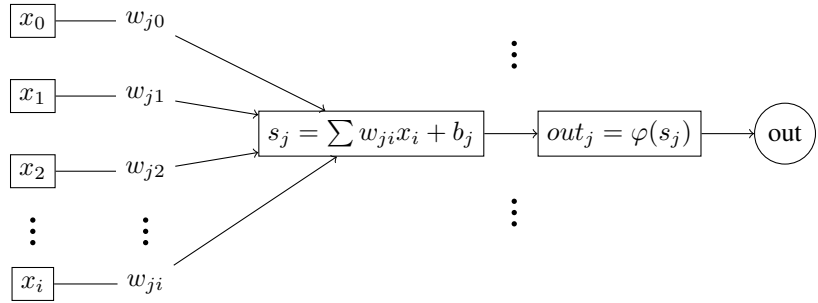

An artificial neuron is a simplified model of a biological neuron. The principle of its operation is shown in the figure~\ref{simple-neuron}. First, the neuron receives an n-dimensional vector of input values $ X=(x_{1},...,x_{n})$ and a vector of weights $W=(w_{1},...,w_{n})$, denoting ``strengthening'' of inter-neuronal connections. The sum of the products of the input values and the weights $ s_j $ is calculated . Then the \hyperref[sec:activation]{activation function} $\varphi$ is applied to the result . Additionally, the amount of offset $b_j $ .
Many neurons form layers. The layers, in turn, form a neural network. The input layer receives data, processes and transmits it to the neurons of the hidden layer. Each subsequent layer works in the same way until the output.
\tikzset{%
every neuron/.style={
  circle,
  draw,
  minimum size=1cm
},
neuron missing/.style={
  draw=none,
  scale=4,
  text height=0.333cm,
  execute at begin node=\color{black}$\vdots$
},
}
\begin{figure}[h]
  \centering
\begin{tikzpicture}[x=1.5cm, y=1.5cm, >=stealth]

\foreach \m/\l [count=\y] in {1,2,3,missing,4}
\node [every neuron/.try, neuron \m/.try] (input-\m) at (0,2.5-\y) {};

\foreach \m [count=\y] in {1,missing,2}
\node [every neuron/.try, neuron \m/.try ] (hidden-\m) at (2,2-\y*1.25) {};

\foreach \m [count=\y] in {1,missing,2}
\node [every neuron/.try, neuron \m/.try ] (output-\m) at (4,1.5-\y) {};

\foreach \l [count=\i] in {1,2,3,n}
\draw [<-] (input-\i) -- ++(-1,0)
  node [above, midway] {$I_\l$};

\foreach \l [count=\i] in {1,k}
\node [above] at (hidden-\i.north) {$H_\l$};

\foreach \l [count=\i] in {1,p}
\draw [->] (output-\i) -- ++(1,0)
  node [above, midway] {$O_\l$};

\foreach \i in {1,...,4}
\foreach \j in {1,...,2}
  \draw [->] (input-\i) -- (hidden-\j);

\foreach \i in {1,...,2}
\foreach \j in {1,...,2}
  \draw [->] (hidden-\i) -- (output-\j);

\foreach \l [count=\x from 0] in {Input, Hidden, Output}
\node [align=center, above] at (\x*2,2) {\l \\ layer};
\end{tikzpicture}
\caption{Simple neural network diagram} \label{simple-network}
\end{figure}
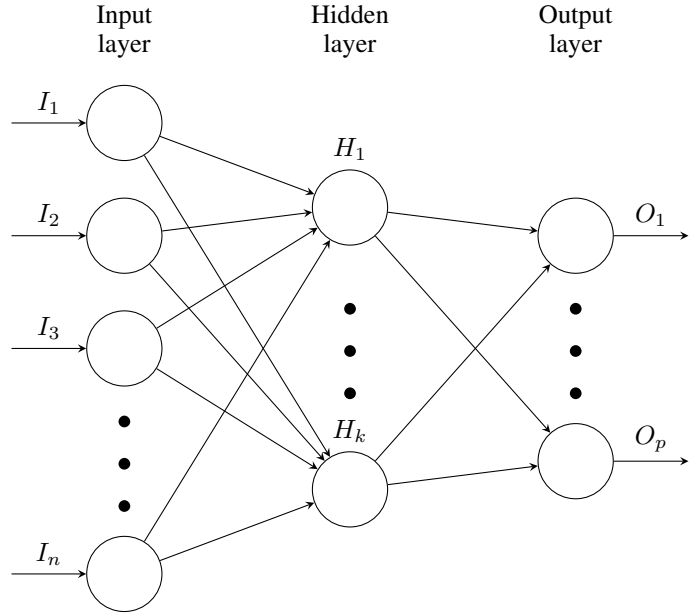
A neural network with a large number of hidden layers is called deep. The field of machine learning that uses deep neural networks is called deep learning.
\subsection{Activation function}
\label{sec: activation}
The weighted sum of the inputs is a linear combination, which means that regardless of the number of layers, the values of the output layer depend only on the inputs of the first layer.
The activation function of the neuron ensures the normalization of the calculated sum and the nonlinearity of the neural network. Many neural network models also require the activation function to be monotonic and continuously differentiable over the entire domain.

There are many activation functions. The most common ones are shown in the table \ref{actvs} .
\pgfplotsset{
  every axis plot/.append style={
    line width=3pt
  }
}
\begin{table}[h]
  \centering
  \caption{Popular activation functions}\label{actvs}
  \begin{tabular}{|c|c|c|}
    \hline
    \hyperlink{name}{Name} & \hyperlink{func}{Function} & \hyperlink{chart}{Type}\\
    \hline
    sigmoid function  &  \resizebox{0.2\hsize}{!}{$\sigma(x)=\frac{1}{1+e^{-x}}$}
    &
    \begin{tikzpicture}[baseline={(0,0.8)}, scale=0.6]
      \begin{axis}[
        axis equal image,
        axis lines=middle,
        axis line style={->},
        x label style={at={(axis description cs:0.5,-0.1)},anchor=north},
        y label style={at={(axis description cs:-0.1,.5)},rotate=90,      anchor=south},
        extra x ticks=0,
        ymin=0,ymax=1,
        ytick={0, 1},
        xtick={-1, 0, 1}
        ]
        \addplot[domain=-1:1, variable=\x] ({\x},{1/(1+exp(-4*\x))});
      \end{axis}
    \end{tikzpicture}

    \\
    \hline
    Hyperbolic tangent
    &
    \resizebox{0.2\hsize}{!}{$f(x)=\frac{e^x-e^{-x}}{e^x+e^{-x}}$}
    &
    \begin{tikzpicture}[baseline={(0,1.5)},scale=0.6]
      \begin{axis}[
        axis equal image,
        axis lines=middle,
        axis line style={->},
        x label style={at={(axis description cs:0.5,-0.1)},anchor=north},
        y label style={at={(axis description cs:-0.1,.5)},rotate=90,      anchor=south},
         extra x ticks=0,
        ymin=-1,ymax=1,
        ytick={-1, 0, 1},
        xtick={-1, 0, 1}
        ]
        \addplot[domain=-1:1, variable=\x]({\x},{tanh(4*\x)});
      \end{axis}
    \end{tikzpicture}
    \\
    \hline
    ReLU & \resizebox{0.2\hsize}{!}{$f(x) =\begin{cases}
    0, & x<0 \\
    x, & x \geq 0.
    \end{cases}$} &
    \begin{tikzpicture}[baseline={(0,0.8)},scale=0.6]
      \begin{axis}[
        axis equal image,
        axis lines=middle,
        axis line style={->},
        x label style={at={(axis description cs:0.5,-0.1)},anchor=north},
        y label style={at={(axis description cs:-0.1,.5)},rotate=90,anchor=south},
         extra x ticks=0,
        ymin=0,ymax=1,
        ytick={0, 1},
        xtick={-1, 0, 1}
        ]
        \addplot[domain=-1:1, variable=\x]({\x},{ifthenelse(\x<0,0,\x)});
      \end{axis}
    \end{tikzpicture}
    \\
    \hline
  \end{tabular}
\end{table}

It should also necessary to mention the Softmax function. This feature is often used on the last layer of deep neural networks in classification problems. Let the last layer of the network contain N neurons, each of which corresponds to a certain class. Then the value of the output of the i-th neuron is calculated by the formula:
\begin{equation}\label{1}
   y_i=\frac{e^{z_i}}{\sum\limits_{j=1}^{N}e^{z_j}}
\end{equation}

Thus, the result of each neuron will take values from the range $[0,1]$, and their sum is 1. As a result, the network will give the probabilities of the ratio of the input data to the given classes.
\subsection{Training neural networks}
The training of neural networks means the selection of the values of the weights of the connections for the effective solution of the task. Initially, weights are set randomly. Then, in the process of running the test data through the network, the weights are adjusted so that in the end the network gives the correct answers.
The learning process is cyclical. During one iteration, a packet is fed to the network containing a number of elements from the input data. A single pass through the network of the entire set of test data is called an epoch.

In order to control the learning process, it is necessary to somehow evaluate the work of the network. For this, a loss function (cost function) is introduced, which calculates~the difference between corrected~and obtained results and forms a certain numerical value characterizing the magnitude of the network operation error. Thus, the task of training the network is reduced to the task of finding the global minimum of a given function. The table~\ref{loss_funcs} contains the most frequently used loss functions, where $y_i$ is the expected value of the i-th neuron, $x_i$ is the obtained value of the i-th neuron, n is the number of output neurons.
\begin{table}[h]
  \centering
  \caption{Popular loss functions} \label{loss_funcs}
  \begin{tabular}{|c|c|}
    \hline
    \hyperlink{name}{Name} & \hyperlink{func}{Function}\\
    \hline
   Mean square error & $E=\frac{1}{N}\displaystyle\sum\limits_{i=1}^{n}(y_i - x_i)^2$\\
    \hline
     Average absolute error & $E=\frac{1}{N}\displaystyle\sum\limits_{i=1}^{n}|y_i - x_i|$ \\
    \hline
    Upper bound & $E=\frac{1}{N}\displaystyle\sum\limits_{i=1}^{n}\max(1-x_i, y_i, 0)$ \\
    \hline
    Categorical Cross Entropy & $E=-\displaystyle\sum\limits_{i=1}^{n}(x_i \cdot log(y_i))$ \\
    \hline
  \end{tabular}
\end{table}
One of the popular methods in training deep neural networks is the backpropagation algorithm.

Let the network have L layers, $ a^l$ , $w_{}^l$ , $ b^l $ - vectors of values, weights and displacements of neurons on the $ l $ -th layer .. There are also N training pairs (x, y).
In the learning process, the following iterations occur in cycles:

{\begin{enumerate}
    \item A vector x from the training set is fed to the network input, for each layer calculate the net input values~$\sigma(z^l)$:
    \begin{equation}\label{1.2}
      \sigma(z^l) = w^la^{l-1}+b^l a^l
    \end{equation}
    \item Calculate the value of the cost function~$C$:
    \begin{equation}\label{1.3}
    C = \frac{1}{2}\sum_j{(y_j-a_j^L)^2}
  \end{equation}
    \item Calculate the error values~$\delta_j^L$~of the output layer:
    \begin{equation}\label{1.4}
       \delta_j^L=\frac{\delta C}{\delta a_j^L}\sigma'(z_j^L)
    \end{equation}
    \item Calculate errors~$\delta_j^l$~for each previous layer:
    \begin{equation}\label{1.5}
      \delta_j^l=\sum_k{w_{kj}^{l+1} \delta_k^{l+1}\sigma'(z_j^l)}
    \end{equation}
    \item Calculate the gradient of the cost function~$\frac{\delta C}{\delta w_{jk}^l}$~:
    \begin{equation}\label{1.6}
      \frac{\delta C}{\delta w_{jk}^l} = a_k^{l-1} \delta_j^l
    \end{equation}
    \item Update link weights~$ w_{ij}^l$~:
    \begin{equation}\label{1.7}
       w_{ij}^l=w_{ij}^l-\mu\frac{\delta C}{\delta w_{jk}^l},\hspace{1em}<\mu \leqslant 1
    \end{equation}
\end{enumerate}
\vspace {1em}
In addition to the method described above, other algorithms are often used for training, for example, RMSprop and Adam optimizers~\cite{zou2019sufficient}. Root Mean Square Propagation (RMSProp) that also maintains per-parameter learning rates that are adapted based on the average of recent magnitudes of the gradients for the weight. The RMSprop optimizer limits the uncertainties in the vertical direction. Hence, the learning rate can be increased, and the deep learning model could take more massive steps in the horizontal direction, gathering quicker. Whereas, Adam is a popular algorithm in the field of deep learning because it achieves good results fast. In Adam optimizer, a learning rate is maintained for each network parameter and separately adapted as learning unfolds.

\vspace {1em}
These methods belong to the \textbf{ supervised learning} algorithms , the most common type of learning, in which the network learns from pre-labeled data where the correct answers are already known.

There are other approaches to training neural networks:

\textbf{Reinforcement learning} is a method that assumes the presence of some environment in which the network operates. Such an environment reacts to the actions of the model and gives it certain signals.

\textbf{Unsupervised learning} - learning in which the network does not have the correct answers in advance and independently searches for common and distinctive features of the input data.

\textbf{Genetic algorithms} - algorithms that mimic the evolutionary mechanisms of the development of a biological population, act as an alternative to the error backpropagation algorithm. The value of an arbitrary weighting factor in a neural network is called a gene. Genes form chromosomes, and chromosomes form a population. Further, within one epoch, with certain probabilities occurs:
 \begin{itemize}
     \item Crossing of chromosomes - the formation of a new chromosome from the genes of the other two
     \item mutation - random change of an arbitrary gene
     \item adaptation - the chromosomes showing the worst results are eliminated from the population.
 \end{itemize}

\subsection{Problems of training deep neural networks}

In learning algorithms based on the backpropagation method. The error value depends on the derivative of the activation function, so when using the sigmoid activation function, the error value decreases very quickly when propagating from the last layer to the first, thus the weights in the early layers are poorly corrected. Similarly,~the uncontrolled gradient problems can occur when the error~value becomes very large.
 A simple way to solve this problem is to use the ReLU function, whose derivative takes the values either 0 or 1.
To solve such problems,  preprocessed input data can be resorted, it is often recommended to limit the input data to the range $[0; 1]$. For example, in images, values can range from 0 to 255 and for better learning stability they can be divided by 255. As another optimization method,  values can be centered to calculate the average value for the input image and subtract it from each pixel, so the most average value of the data in the image will be equal to 0. Together with this method, the standard deviation is often normalized, setting its value to 1.

Network retraining is a problem when the network learns to analyze objects well only from the training set and does not work well with new data. One of the methods for solving this problem is Dropout, the essence of which is as follows: At each training iteration, neurons with some probability are turned off. The remaining neurons are trained by the backpropagation method of oshiyuki, after which the neurons return to the network.

\begin{figure}[h]
    \centering
    \begin{tikzpicture}

        \node[circle, draw, thick] (i1) {};
        \node[circle, draw, thick, above=1em of i1] (i2) {};
        \node[circle, draw, thick, above=1em of i2] (i3) {};
        \node[circle, draw, thick, below=1em of i1] (i4) {};
        \node[circle, draw, thick, below=1em of i4] (i5) {};

        \node[circle, draw, thick, right=2em of i1] (h1) {};
        \node[circle, draw, thick, right=2em of i2] (h2) {};
        \node[circle, draw, thick, right=2em of i3] (h3) {};
        \node[circle, draw, thick, right=2em of i4] (h4) {};
        \node[circle, draw, thick, right=2em of i5] (h5) {};

        \node[circle, draw, thick, right=2em of h1] (hh1) {};
        \node[circle, draw, thick, right=2em of h2] (hh2) {};
        \node[circle, draw, thick, right=2em of h3] (hh3) {};
        \node[circle, draw, thick, right=2em of h4] (hh4) {};
        \node[circle, draw, thick, right=2em of h5] (hh5) {};

        \node[circle, draw, thick, right=2em of hh2] (o1) {};
        \node[circle, draw, thick, right=2em of hh4] (o2) {};

        \draw[-stealth, thick] (i1) -- (h1);
        \draw[-stealth, thick] (i1) -- (h2);
        \draw[-stealth, thick] (i1) -- (h3);
        \draw[-stealth, thick] (i1) -- (h4);
        \draw[-stealth, thick] (i1) -- (h5);
        \draw[-stealth, thick] (i2) -- (h1);
        \draw[-stealth, thick] (i2) -- (h2);
        \draw[-stealth, thick] (i2) -- (h3);
        \draw[-stealth, thick] (i2) -- (h4);
        \draw[-stealth, thick] (i2) -- (h5);
        \draw[-stealth, thick] (i3) -- (h1);
        \draw[-stealth, thick] (i3) -- (h2);
        \draw[-stealth, thick] (i3) -- (h3);
        \draw[-stealth, thick] (i3) -- (h4);
        \draw[-stealth, thick] (i3) -- (h5);
        \draw[-stealth, thick] (i4) -- (h1);
        \draw[-stealth, thick] (i4) -- (h2);
        \draw[-stealth, thick] (i4) -- (h3);
        \draw[-stealth, thick] (i4) -- (h4);
        \draw[-stealth, thick] (i4) -- (h5);
        \draw[-stealth, thick] (i5) -- (h1);
        \draw[-stealth, thick] (i5) -- (h2);
        \draw[-stealth, thick] (i5) -- (h3);
        \draw[-stealth, thick] (i5) -- (h4);
        \draw[-stealth, thick] (i5) -- (h5);

        \draw[-stealth, thick] (h1) -- (hh1);
        \draw[-stealth, thick] (h1) -- (hh2);
        \draw[-stealth, thick] (h1) -- (hh3);
        \draw[-stealth, thick] (h1) -- (hh4);
        \draw[-stealth, thick] (h1) -- (hh5);
        \draw[-stealth, thick] (h2) -- (hh1);
        \draw[-stealth, thick] (h2) -- (hh2);
        \draw[-stealth, thick] (h2) -- (hh3);
        \draw[-stealth, thick] (h2) -- (hh4);
        \draw[-stealth, thick] (h2) -- (hh5);
        \draw[-stealth, thick] (h3) -- (hh1);
        \draw[-stealth, thick] (h3) -- (hh2);
        \draw[-stealth, thick] (h3) -- (hh3);
        \draw[-stealth, thick] (h3) -- (hh4);
        \draw[-stealth, thick] (h3) -- (hh5);
        \draw[-stealth, thick] (h4) -- (hh1);
        \draw[-stealth, thick] (h4) -- (hh2);
        \draw[-stealth, thick] (h4) -- (hh3);
        \draw[-stealth, thick] (h4) -- (hh4);
        \draw[-stealth, thick] (h4) -- (hh5);
        \draw[-stealth, thick] (h5) -- (hh1);
        \draw[-stealth, thick] (h5) -- (hh2);
        \draw[-stealth, thick] (h5) -- (hh3);
        \draw[-stealth, thick] (h5) -- (hh4);
        \draw[-stealth, thick] (h5) -- (hh5);

        \draw[-stealth, thick] (hh1) -- (o1);
        \draw[-stealth, thick] (hh1) -- (o2);
        \draw[-stealth, thick] (hh2) -- (o1);
        \draw[-stealth, thick] (hh2) -- (o2);
        \draw[-stealth, thick] (hh3) -- (o1);
        \draw[-stealth, thick] (hh3) -- (o2);
        \draw[-stealth, thick] (hh4) -- (o1);
        \draw[-stealth, thick] (hh4) -- (o2);
        \draw[-stealth, thick] (hh5) -- (o1);
        \draw[-stealth, thick] (hh5) -- (o2);

        \draw[->, thick] (4.5,0) -- node[above] {} (5.5, 0);


        \node[circle, draw, thick, red, fill=red!10, right=8em of hh1] (i1) {};
        \node[circle, draw, thick, red, fill=red!10, above=1em of i1] (i2) {};
        \node[circle, draw, thick, above=1em of i2] (i3) {};
        \node[circle, draw, thick, below=1em of i1] (i4) {};
        \node[circle, draw, thick, below=1em of i4] (i5) {};


        \node[circle, draw, thick, red, fill=red!10, right=2em of i1] (h1) {};
        \node[circle, draw, thick, right=2em of i2] (h2) {};
        \node[circle, draw, thick, red, fill=red!10, right=2em of i3] (h3) {};
        \node[circle, draw, thick, red, fill=red!10, right=2em of i4] (h4) {};
        \node[circle, draw, thick, right=2em of i5] (h5) {};


        \node[circle, draw, thick, right=2em of h1] (hh1) {};
        \node[circle, draw, thick, red, fill=red!10, right=2em of h2] (hh2) {};
        \node[circle, draw, thick, right=2em of h3] (hh3) {};
        \node[circle, draw, thick, red, fill=red!10, right=2em of h4] (hh4) {};
        \node[circle, draw, thick, right=2em of h5] (hh5) {};


        \node[circle, draw, thick, right=2em of hh2] (o1) {};
        \node[circle, draw, thick, right=2em of hh4] (o2) {};

        \draw[-stealth, thick] (i3) -- (h2);
        \draw[-stealth, thick] (i3) -- (h5);
        \draw[-stealth, thick] (i4) -- (h2);
        \draw[-stealth, thick] (i4) -- (h5);
        \draw[-stealth, thick] (i5) -- (h2);
        \draw[-stealth, thick] (i5) -- (h5);

        \draw[-stealth, thick] (h2) -- (hh1);
        \draw[-stealth, thick] (h2) -- (hh3);
        \draw[-stealth, thick] (h2) -- (hh5);
        \draw[-stealth, thick] (h5) -- (hh1);
        \draw[-stealth, thick] (h5) -- (hh3);
        \draw[-stealth, thick] (h5) -- (hh5);

        \draw[-stealth, thick] (hh1) -- (o1);
        \draw[-stealth, thick] (hh1) -- (o2);
        \draw[-stealth, thick] (hh3) -- (o1);
        \draw[-stealth, thick] (hh3) -- (o2);
        \draw[-stealth, thick] (hh5) -- (o1);
        \draw[-stealth, thick] (hh5) -- (o2);
    \end{tikzpicture}
    \caption{Dropout} \label{dropout}
\end{figure}

\subsection{Convolutional Neural Networks} \label{CNN}

\subsubsection{Architecture}
Most of the modern neural networks aimed at image analysis are based on the architecture of the convolutional neural network.
Early neural networks consisted of fully connected layers-layers in which each neuron is connected to each neuron of the next layer, which significantly increased the computational complexity of the system as the number of neurons increased.
Typical convolutional neural networks primarily use convolutional layers.
Convolutional layers are characterized by the use of weight matrices, called filters or kernels, that are smaller than the original data. Such a kernel with a certain step goes through the set of input data $(I)$ and calculates the sums of the products of the corresponding values of the cells and weights, forming a feature map $(I * K)$ One.  convolutional layer can contain several kernels and, accordingly, several feature maps.
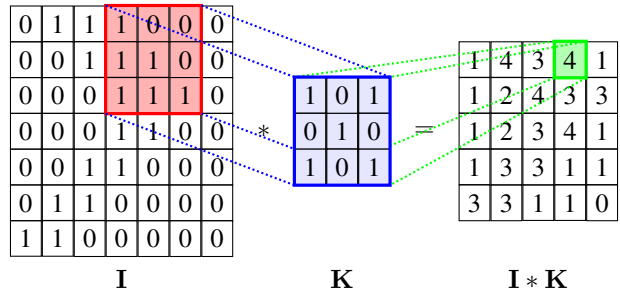
\begin{figure}[h]
\centering
\begin{tikzpicture}

	\matrix (mtr) [matrix of nodes,row sep=-\pgflinewidth, nodes={draw}]
	{
		0 & 1 & 1 & |[fill=red!30]| 1 & |[fill=red!30]| 0 & |[fill=red!30]| 0 & 0\\
		0 & 0 & 1 & |[fill=red!30]| 1 & |[fill=red!30]| 1 & |[fill=red!30]| 0 & 0\\
		0 & 0 & 0 & |[fill=red!30]| 1 & |[fill=red!30]| 1 & |[fill=red!30]| 1 & 0\\
		0 & 0 & 0 & 1 & 1 & 0 & 0\\
		0 & 0 & 1 & 1 & 0 & 0 & 0\\
		0 & 1 & 1 & 0 & 0 & 0 & 0\\
		1 & 1 & 0 & 0 & 0 & 0 & 0\\
	};

	\draw[very thick, red] (mtr-1-4.north west) rectangle (mtr-3-6.south east);

	\node [below= of mtr-5-4.south] (lm) {$\bf I$};

	\node[right = 0.2em of mtr] (str) {$*$};

	\matrix (K) [right=0.2em of str,matrix of nodes,row sep=-\pgflinewidth, nodes={draw, fill=blue!30}]
	{
		1 & 0 & 1 \\
		0 & 1 & 0 \\
		1 & 0 & 1 \\
	};
	\node [below = of K-3-2.south] (lk) {$\bf K$};

	\node [right = 0.2em of K] (eq) {$=$};

	\matrix (ret) [right=0.2em of eq,matrix of nodes,row sep=-\pgflinewidth, nodes={draw}]
	{
		1 & 4 & 3 & |[fill=green!30]| 4 & 1\\
		1 & 2 & 4 & 3 & 3\\
		1 & 2 & 3 & 4 & 1\\
		1 & 3 & 3 & 1 & 1\\
		3 & 3 & 1 & 1 & 0\\
	};
	\node [below = of ret-4-3.south] (lim) {${\bf I} * {\bf K}$};

	\draw[very thick, green] (ret-1-4.north west) rectangle (ret-1-4.south east);

	\draw[densely dotted, blue, thick] (mtr-1-4.north west) -- (K-1-1.north west);
	\draw[densely dotted, blue, thick] (mtr-3-4.south west) -- (K-3-1.south west);
	\draw[densely dotted, blue, thick] (mtr-1-6.north east) -- (K-1-3.north east);
	\draw[densely dotted, blue, thick] (mtr-3-6.south east) -- (K-3-3.south east);

	\draw[densely dotted, green, thick] (ret-1-4.north west) -- (K-1-1.north west);
	\draw[densely dotted, green, thick] (ret-1-4.south west) -- (K-3-1.south west);
	\draw[densely dotted, green, thick] (ret-1-4.north east) -- (K-1-3.north east);
	\draw[densely dotted, green, thick] (ret-1-4.south east) -- (K-3-3.south east);

	\matrix (K) [right=0.2em of str,matrix of nodes,row sep=-\pgflinewidth, nodes={draw, fill=blue!10}]
	{
		1 & 0 & 1 \\
		0 & 1 & 0 \\
		1 & 0 & 1 \\
	};

	\draw[very thick, blue] (K-1-1.north west) rectangle (K-3-3.south east);

\end{tikzpicture}
\caption{Convolutional Neural Network} \label{convolution}
\end{figure}

Since the features have already been detected, to simplify further calculations, it can be reduced the granularity of the input data. This provides a downsampling (pooling) layer, reducing the dimension of the input feature maps: from several neighboring neurons, the maximum or average value is taken, thereby forming a neuron of the feature map of a lower dimension. This reduces the number of parameters used in further network calculations.

\begin{figure}[h]
    \centering
    \begin{tikzpicture}
        \draw[step=1cm,very thin] (-2,-2) grid (2,2);
        \fill[blue, opacity=0.5] (0,2) rectangle (-2,0);
        \draw[thick,->] (3,0) -- (4,0) node[anchor=north west] {};
        \draw[step=1cm,very thin] (5,-1) grid (7,1);
        \fill[blue, opacity=0.5] (5,0) rectangle (6,1);
    \end{tikzpicture}
    \caption{Subsampling} \label{pooling}
\end{figure}
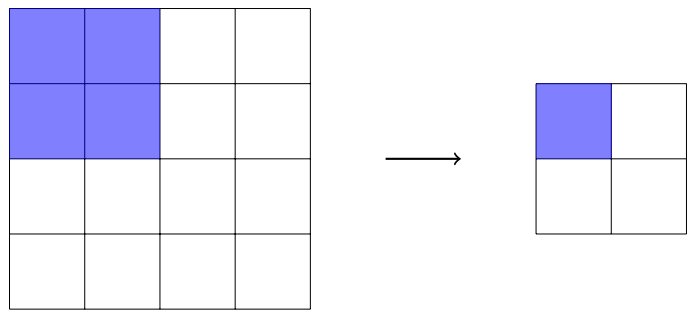

A convolutional neural network can have multiple pairs of alternating convolutional and downsampling layers.
Thus, using the example of images, on the initial layers, the network finds such simple features as borders and corners.Then more and more complicated structures are defined as we go deeper into the network: from the simple forms to the whole categories irrespective of where they are located. The network ends with completely connected standard layers which correspond to the resulting characteristics in the class.
\subsubsection{VGG}
The VGG architecture was proposed in 2014 by~\cite{simonyan2014convolutional}. The main feature of the network is the use of consecutive convolutional layers with 3x3 filters instead of the previously used convolutional layers with large filters 5x5, 7x7, 11x11. This made it possible to reduce the number of network parameters while maintaining efficiency.

The table \ref{vgg} lists the various VGG configurations, the most famous of which are VGG-16 (D) and VGG-19 (E), named by the number of layers containing weights. Maxpool is a downsampling layer with a 2x2 max function. FC is a fully connected layer. All hidden layers use the ReLU activation function.

\begin{table}[h]
    \centering
    \caption{VGG Configs} \label{vgg}
    \resizebox{0.8\textwidth}{!}{%
        \begin{tabular}{|c|c|c|c|c|c|} \hline
        A & A-LRN & B & C & D & E \\ \hline
        \multicolumn{6}{|c|}{input ($224 \times 224$ RGB)} \\ \hline
        conv3-64 & conv3-64 & conv3-64 & conv3-64 & conv3-64 & conv3-64 \\
        & \textbf{LRN} & \textbf{conv3-64} & conv3-64 & conv3-64 & conv3-64\\ \hline
        \multicolumn{6}{|c|}{Maxpool} \\ \hline
        conv3-128 & conv3-128 & conv3-128 & conv3-128 & conv3-128 & conv3-128 \\
        & & \textbf{conv3-128} & conv3-128 & conv3-128 & conv3-128 \\ \hline
        \multicolumn{6}{|c|}{Maxpool} \\ \hline
        conv3-256 & conv3-256 & conv3-256 & conv3-256 & conv3-256 & conv3-256 \\
        conv3-256 & conv3-256 & conv3-256 & conv3-256 & conv3-256 & conv3-256 \\
        & & & \textbf{conv1-256} & \textbf{conv3-256} & conv3-256 \\
        & & & & & \textbf{conv3-256} \\ \hline
        \multicolumn{6}{|c|}{Maxpool} \\ \hline
        conv3-512 & conv3-512 & conv3-512 & conv3-512 & conv3-512 & conv3-512 \\
        conv3-512 & conv3-512 & conv3-512 & conv3-512 & conv3-512 & conv3-512 \\
        & & & \textbf{conv1-512} & \textbf{conv3-512} & conv3-512 \\
        & & & & & \textbf{conv3-512} \\ \hline
        \multicolumn{6}{|c|}{Maxpool} \\ \hline
        conv3-512 & conv3-512 & conv3-512 & conv3-512 & conv3-512 & conv3-512 \\
        conv3-512 & conv3-512 & conv3-512 & conv3-512 & conv3-512 & conv3-512 \\
        & & & \textbf{conv1-512} & \textbf{conv3-512} & conv3-512 \\
        & & & & & \textbf{conv3-512} \\ \hline
        \multicolumn{6}{|c|}{Maxpool} \\ \hline
        \multicolumn{6}{|c|}{FC-4096} \\ \hline
        \multicolumn{6}{|c|}{FC-4096} \\ \hline
        \multicolumn{6}{|c|}{FC-1000} \\ \hline
        \multicolumn{6}{|c|}{Softmax} \\ \hline
        \end{tabular}}
        \end{table}

\subsubsection{Inception}
This model \cite{1512.00567}, developed by Google, in 2014 took 1st place in the annual competition for image classification-ILSVRC. A key innovation of this network was the use of nested modules as layers, which are a set of filters of different dimensions, with the subsequent merging of their results.

\begin{figure}[h]
    \centering
    \begin{tikzpicture}
        [align=center]
        \tikzstyle{rectangle_style}=[rectangle, draw, minimum height = 15mm, minimum width = 10mm]
        \tikzstyle{arrow} = [thick,->,>=stealth]
        \node (in) at (0,0) [rectangle_style] {Previous layer};
        \node (cv1_1) at (-1.5,-2.5) [rectangle_style] {Conv 1x1};
        \node (cv3_1) at (-1.5,-5) [rectangle_style] {Conv 3x3};
        \node (cv1_2) at (2.0, -2.5) [rectangle_style] {Conv 1x1};
        \node (cv5_1) at (2.0,-5) [rectangle_style] {Conv 5x5};
        \node (mp_1)  at (5,-2.5) [rectangle_style] {Maxpool\\3x3};
        \node (cv1_3) at (5,-5) [rectangle_style] {Conv 1x1};
        \node (cv1_0) at (-5.5, -3.75) [rectangle_style] {Conv 1x1};
        \node (out) at (0,-7.5) [rectangle_style] {Combine filters};

        \draw [arrow] (in) -- (cv1_1);
        \draw [arrow] (in) -- (cv1_2);
        \draw [arrow] (in) -- (cv1_0);
        \draw [arrow] (in) -- (mp_1);
        \draw [arrow] (cv1_1) -- (cv3_1);
        \draw [arrow] (cv1_2) -- (cv5_1);
        \draw [arrow] (mp_1) -- (cv1_3);
        \draw [arrow] (cv1_3) -- (out);
        \draw [arrow] (cv5_1) -- (out);
        \draw [arrow] (cv3_1) -- (out);
        \draw [arrow] (cv1_0) -- (out);

    \end{tikzpicture}
    \caption{Inception Module} \label{inception-module}
\end{figure}
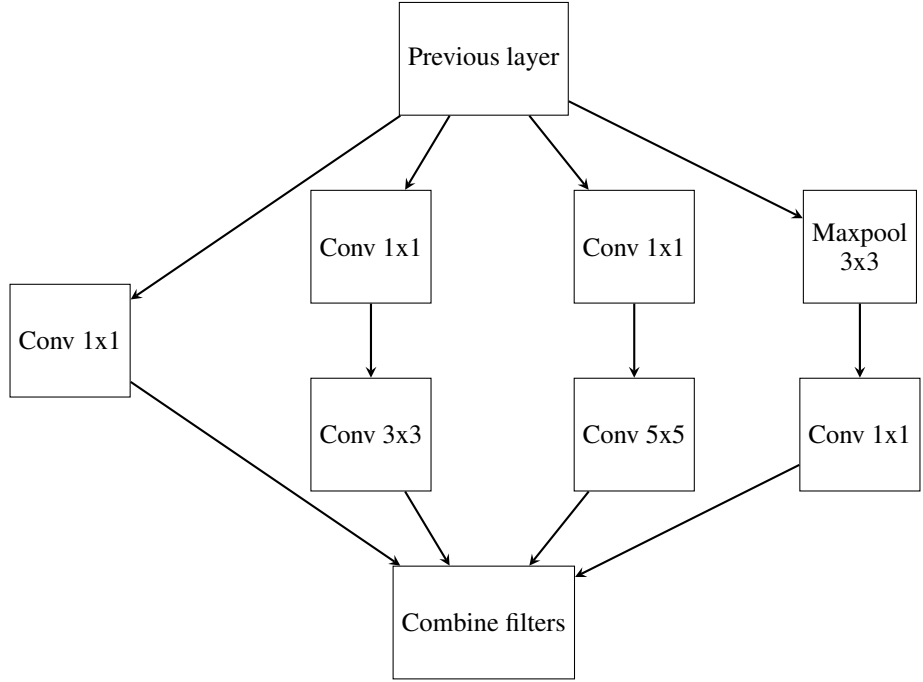

Also, Inception completely abandoned the use of fully connected layers, instead of them, a global average pooling is used, which converts each feature map to one number, forming a vector of averaged values. This innovation made it possible to significantly reduce the number of parameters and, as a consequence, the computational complexity of the network. Later, improved versions of Inception were developed, in which the 5x5 layer was replaced with two successive 3x3 layers, and all layers with $N\times N$ filters were replaced with a $1\times N$ and $N\times 1$ filter stack, which also reduced the number of parameters.

\subsubsection{ResNet}
ResNet~\cite{ResNet}, also known as residual neural network, won the ILSVRC in $2015$. Its feature was the presence of transmission connections that transmit information unchanged to deeper parts of the network, this information is summed up with the value calculated on the missing layers and transmitted further. The block shown in Fig.~\ref{res-net} demonstrates the building block of such a network.

\begin{figure}[h]
\centering
    \begin{tikzpicture}
        \tikzset{
            connector/.style={
            -latex,
            font=\scriptsize
            },
            rectangle connector/.style={
                connector,
                to path={(\tikztostart) -- ++(#1,0pt) \tikztonodes |- (\tikztotarget) },
                pos=0.5
            },
            rectangle connector/.default=-2cm,
            straight connector/.style={
                connector,
                to path=--(\tikztotarget) \tikztonodes
            }
        }
        [align=center]
        \tikzstyle{rectangle_style}=[rectangle, draw, minimum height = 15mm, minimum width = 10mm]
        \tikzstyle{arrow} = [thick,->,>=stealth]
        \node (in) at (0,.3) [] {};
        \node (midin) at (0,-0.6) [] {};
        \node (cv3_1) at (0,-2) [rectangle_style] {Conv 3x3};
        \node (relu2) at (-0.8,-3.5) [] {$ReLU$};
        \node (cv3_2) at (0,-5) [rectangle_style] {Conv 3x3};
        \node (plus) at (0,-7) [circle, draw] {+};
        \node (out) at (0,-8.5) [] {};
        \node (x) at (0,.5) [] {$x$};
        \node (x2) at (-2.5,-4) [] {$x$};
        \node (f_x) at (2,-4) [] {$F(x)$};
        \node (H_x) at (1.5,-7) [] {$F(x)+x$};
        \node (relu2) at (- 0.8,-8) [] {$ReLU$};

        \draw [arrow, rectangle connector] (midin) to  node[] {} (plus);
        \draw [arrow] (in) -- (cv3_1);
        \draw [arrow] (cv3_1) -- (cv3_2);
        \draw [arrow] (cv3_2) -- (plus);
        \draw [arrow] (plus) -- (out);

    \end{tikzpicture}
\caption{Residual net block} \label{res-net}
\end{figure}
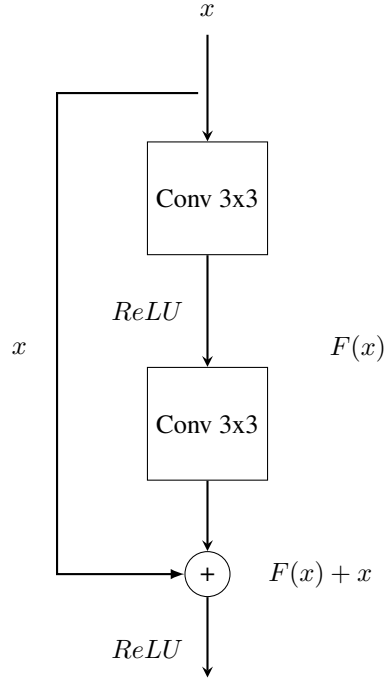

\subsubsection{DenseNet}
DenseNet \cite{DenseNet} is a dense convolutional network, similar to ResNet, but with the difference that all blocks of the network are connected by direct connections, so each block receives information from all previous ones.

\newcommand{\conv}[1]{$\left[\begin{array}{ll} \text{1}\times \text{1} \text{convolution}\\ \text{3}\times \text{3} \text{convolution} \end{array}\right] \times \text{#1}$}
\newcommand{\cross}[1]{#1 $\times$ #1}
\begin{table}[h]
    \centering
    \caption{DenseNet}
    \label{densenets}
    \resizebox{0.845\textwidth}{!}{%
    \begin{tabular}{|c|c|c|l|l|l|}
        \hline
        Layer                                                                        & Output size & DenseNet-121                                                                                                               & \multicolumn{1}{c|}{DenseNet-169}                                                                     & \multicolumn{1}{c|}{DenseNet-201}                                                                     & \multicolumn{1}{c|}{DenseNet-264}                                                                       \\ \hline
        Convolution                                                                     & \cross{112} & \multicolumn{4}{c|}{\cross{7} convolution, step 2}                                                                                                                                                                                                                                                                                                                                                                                                    \\ \hline
        Pooling                                                                         & \cross{56}   & \multicolumn{4}{c|}{\cross{3} maximum pooling, step 2}                                                                                                                                                                                                                                                                                                                                                                                                \\ \hline
        \begin{tabular}[c]{@{}c@{}}Dense block\\ (1)\end{tabular}                       & \cross{56}   & \multicolumn{1}{l|}{\conv{6}}  & \conv{6}  & \conv{6}  & \conv{6} \\ \hline
        \multirow{2}{*}{\begin{tabular}[c]{@{}c@{}}Intermediate layer\\ (1)\end{tabular}} & \cross{56}  & \multicolumn{4}{c|}{\cross{1} convolution}                                                                                                                                                                                                                                                                                                                                                                                                              \\ \cline{2-6}
                                                                                        & \cross{28}   & \multicolumn{4}{c|}{\cross{2} average pooling, step 2}                                                                                                                                                                                                                                                                                                                                                                                             \\ \hline
        \begin{tabular}[c]{@{}c@{}}Dense block\\ (2)\end{tabular}                       & \cross{28}   & \multicolumn{1}{l|}{\conv{12}} & \conv{12}& \conv{12} & \conv{12} \\ \hline
        \multirow{2}{*}{\begin{tabular}[c]{@{}c@{}} Intermediate layer\\ (2)\end{tabular}} & \cross{28}  & \multicolumn{4}{c|}{\cross{1} convolution}                                                                                                                                                                                                                                                                                                                                                                                                              \\ \cline{2-6}
                                                                                        & \cross{14}   & \multicolumn{4}{c|}{\cross{2} average pooling, step 2}                                                                                                                                                                                                                                                                                                                                                                                             \\ \hline
        \begin{tabular}[c]{@{}c@{}}Dense block\\ (3)\end{tabular}                       & \cross{14}   & \multicolumn{1}{l|}{\conv{24}} & \conv{32} & \conv{48} & \conv{64} \\ \hline
        \multirow{2}{*}{\begin{tabular}[c]{@{}c@{}}Intermidiate layer\\ (3)\end{tabular}} & \cross{14}   & \multicolumn{4}{c|}{\cross{1} convolution}                                                                                                                                                                                                                                                                                                                                                                                                              \\ \cline{2-6}
                                                                                        & \cross{7}     & \multicolumn{4}{c|}{\cross{2} average pooling, step 2}                                                                                                                                                                                                                                                                                                                                                                                             \\ \hline
        \begin{tabular}[c]{@{}c@{}}Dense block\\ (4)\end{tabular}                       & \cross{7}   & \multicolumn{1}{l|}{\conv{16}} & \conv{32} & \conv{32} & \conv{48}  \\ \hline
        \multirow{2}{*}{\begin{tabular}[c]{@{}c@{}}classification\\ Layer\end{tabular}} & \cross{1} & \multicolumn{4}{c|}{\cross{7} global average pooling}                                                                                                                                                                                                                                                                                                                                                                                            \\ \cline{2-6}
                                                                                        &                 & \multicolumn{4}{c|}{1000D
fully connected layer, softmax}                                                                                                                                                                                                                                                                                                                                                                                               \\ \hline
    \end{tabular} } \end{table}


\subsubsection{Testing}
All of these models were tested on ImageNet datasets, which include 1000 classes, over 1.3 million training and 50 thousand verification images. Since networks perform the task of classification and use a Softmax layer at the end, the result of the network is the vector $(x_1, x_2, ... x_n)$, where n is the number of classes, and $x_i$ is the probability of the ratio of the input image to the i-th class. The accuracy of the networks was measured in two variants Top-1 and Top-5. Top-1 means that the largest $ x_i $ corresponds to the correct class, and Top-5 means that the correct class belongs to the five highest values in the output vector of the network. The table \ref{conv-test} shows the results of the tests carried out using the accuracy metric-the ratio of the proportion of correct answers to their total number. The number of samples are less in the covid-19 class, hence confidence interval(CI) of the all the performance metric were calculated by using following formula~\ref{eq1}:

\begin{equation}\label{eq1}
r = z\sqrt{acc(1-acc)/N}
\end{equation}
 Where  $z$ is the number of standard deviation also called the CI level, $acc$ is the performance metrics used for the evaluation of performance models given in equation~\ref{1.13}. And the value of the $N$ is the number of sample for the class used. We have used $95\%$ CI and standard deviation as $1$.

\begin{table}[h]
    \centering
    \caption{Model testing results.\\ Values in brackets are 95\% confidence intervals.} \label{conv-test}
    \begin{tabular}{|c|c|c|c|}
      \hline
     Network & Parameters Qty & Top-1(in \%)  & Top-5(in \%) \\
      \hline
      VGG-16       & 138 357 544   & 71.3~[69.2,73.3]	& 90.1~[90.0,91.4]    \\
      \hline
      VGG-19       & 143 667 240     & 71.3~[69.3,73.3]	& 90.0~[89.9.0,91.3]    \\
      \hline
      Inception V3 & 23 851 784          & 77.9~[76.0,79.8]	& 93.7~[93.6,94.7]   \\
      \hline
      ResNet-50 V2 & 25 613 800         & 76.0~[74.1,77.9]	& 93.0~[92.9.0,94.1]    \\
      \hline
      ResNet-101 V2 & 44 675 560	   & 77.2~[75.3,79.1]	& 93.8~[93.7,94.8]   \\
      \hline
      ResNet-152 V2 & 60 380 648	       & 78.0~[76.1,79.8]& 94.2~[94.1,95.2]    \\
      \hline
      DenseNet-121 & 8 062 504           & 75.0~[73.0,76.9]	& 92.3~[92.2,93.4]    \\
      \hline
      DenseNet-169 & 14 307 880          & 76.2~[74.3,78.1]	& 93.2~[93.1,94.3]   \\
      \hline
      DenseNet-201 & 20 242 984          & 77.3~[75.4,79.1]	& 93.6~[93.5,94.6]    \\
      \hline
    \end{tabular}
  \end{table}

As it can be seen in the table: VGG networks have the largest number of parameters, significantly surpassing other models, while demonstrating the lowest accuracy. DenseNet shows the best ratio of accuracy to parameter count. However, the third version of Inception has relatively slightly more parameters and precision. The best result was shown by ResNet-152 of the second version, but it has a significantly larger number of parameters, which affects the network learning time. AlexNet and ResNet-152, both have about 60M parameters but there is about 10\% difference in their top-5 accuracy. But training a ResNet-152 requires a lot of computations (about 10 times more than that of AlexNet) which means more training time and energy required. VGGNet not only has a higher number of parameters as compared to ResNet-152 but also has a decreased accuracy. It takes more time to train a VGGNet with reduced accuracy. Training an AlexNet takes about the same time as training Inception. The memory requirements are 10 times less with improved accuracy (about 9\%)

\section{Methodology}\label{dev}

\subsection{Staging tasks}
As discussed in introduction, X-rays are less sensitive than CT or PCR, it is a much more affordable and quick diagnostic method, which is an essential criterion during a pandemic as discussed in the introduction section. Figure~\ref{xrays} shows the CXR of three types of patients i.e normal vs pneumonia vs COVID-19. Solving the problem of automatic diagnosis of this disease will reduce the burden on doctors and increase the efficiency of their work.

\addthreeimghere{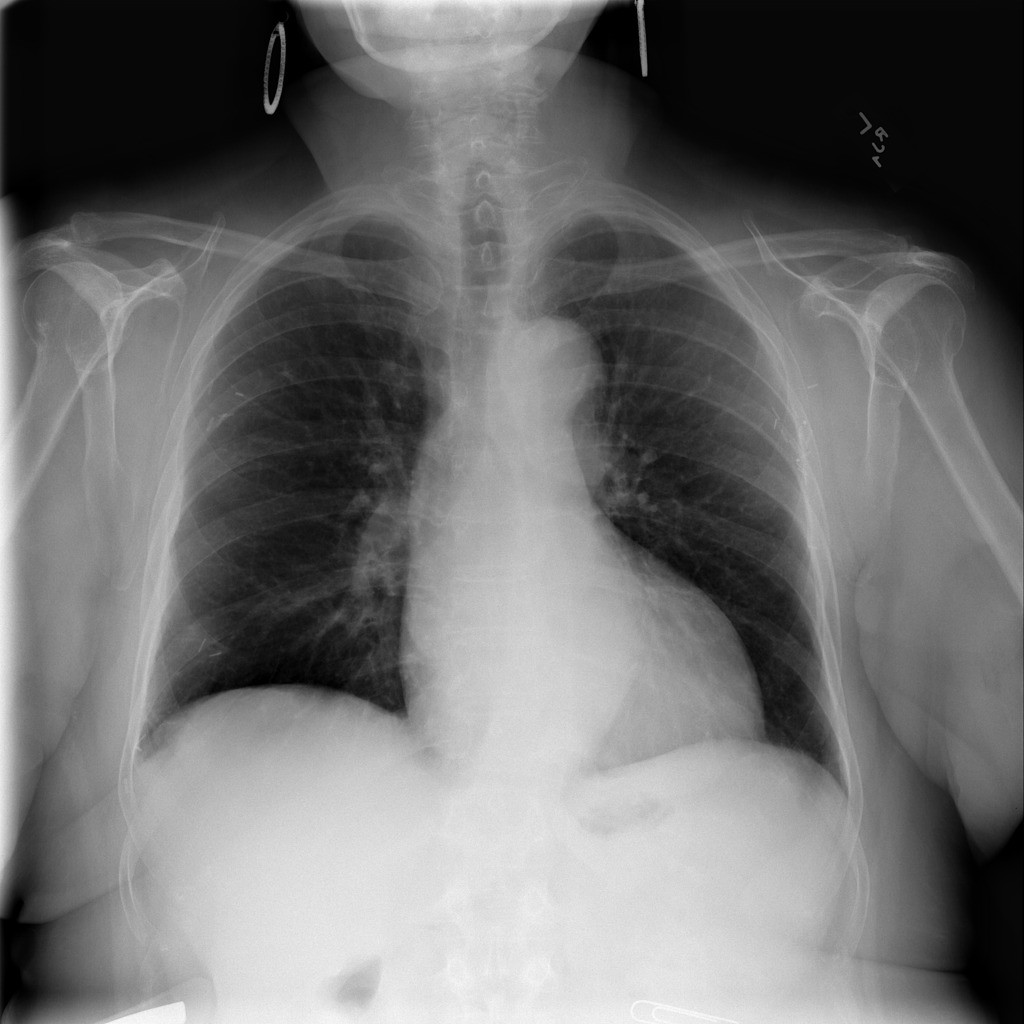}{Normal}{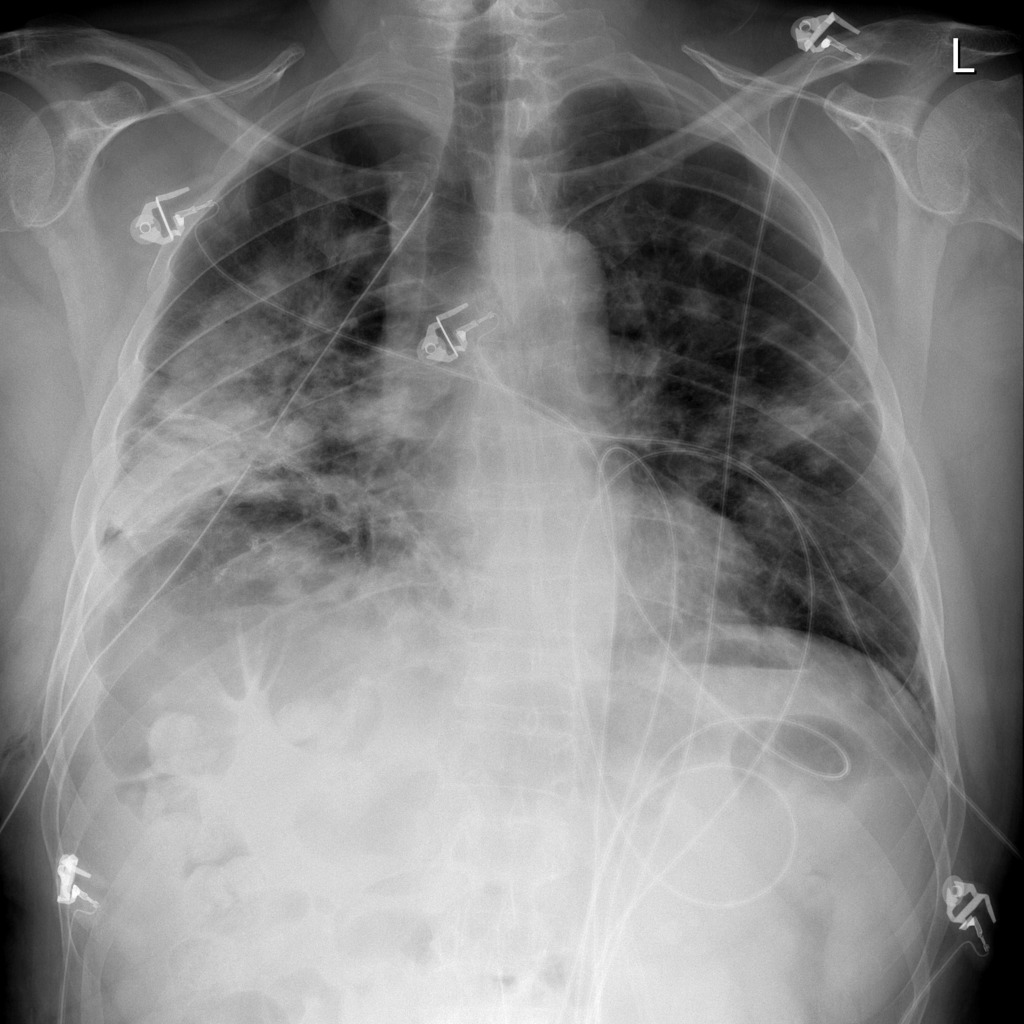}{Pneumonia}{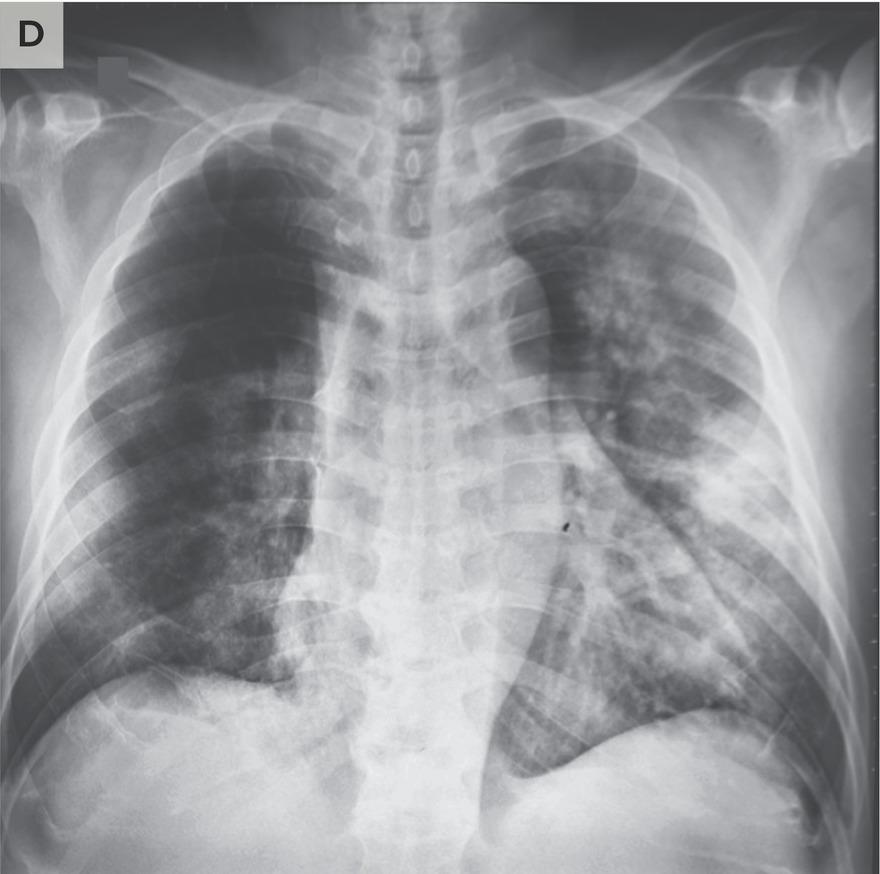}{COVID-19}{Chest X-rays}{xrays}

\subsection{Implementation facilities}
cuDNN is a deep neural network library from Nvidia that allows to use GPU power for calculations.
Python 3 is a flexible and powerful programming language that efficiently performs data analysis and processing tasks.
TensorFlow is a feature-rich open source framework developed by Google that allows to design and train various neural network architectures.
Keras is a high-level deep machine learning API included with TensorFlow.
 \subsection{System configuration}
 All models have been tested on a system with the following specifications:
 \begin{itemize}
    \item Operating system - Ubuntu 20.04
   \item CPU - Intel Core i5 9400F CPU 2.90 Ghz
    \item The amount of RAM - 8 GB
    \item Video card - Nvidia GeForce GTX 1660
 \end{itemize}

\subsection{Quality assessment}
The following metrics were used to assess the quality of the algorithms:
\begin{itemize}
    \item Precision :
     \begin{equation}\label{1.10}
     P = \frac{TP}{TP+FP}
     \end{equation}
    \item Recall:
    \begin{equation}\label{1.11}
    R=\frac{TP}{TP+FN}
    \end{equation}
    \item F1-measure:
    \begin{equation}\label{1.12}
 F1=2 \cdot \frac{P \cdot R}{P+R}
    \end{equation}
     \item Accuracy:
     \begin{equation}\label{1.13}
     acc=\frac{TP+TN}{TP+TN+FP+FN}
     \end{equation}
\end{itemize}
where TP is the number of true positive, FP - false positive, FN - False Negative Answers. \\
Precision denotes the proportion of correctly identified objects of a class relative to all objects assigned to this class. \\ Recall shows the proportion of elements of a class found by the network relative to all elements of this class. \\
F1 combines precision and recall by calculating their harmonic mean.
Also, in the process of training the network, the values of the loss function were considered as a characteristic of the quality of training.

\subsection{Training set}
Collecting data for training neural networks in tasks of this type is a complex and time-consuming process that requires a lot of time and the participation of a large number of people. Therefore, ready-made, already marked datasets were used as a source of training and verification data: ~\cite{kaggle}
A total of 14,197 images were collected, of which 8,066 were healthy patients, 5,558 with pneumonia and 573 with COVID-19. 100 images of each class were selected for training and validation.

The total number of images are shown in the table \ref{inputs}

\begin{table}[h]
    \centering
    \caption{Number of Inputs} \label{inputs}
    \begin{tabular}{|c|c|c|}
      \hline
                & Training & Validation  \\
      \hline
    normal     & 7 966     & 885       \\
      \hline
      Pneumonia & 5 458     & 594       \\
      \hline
      COVID-19  & 473       & 100       \\
      \hline

Total    & 13 897    & 1 579     \\
      \hline
    \end{tabular}
  \end{table}

 Images from the training set have different resolutions, but neural networks require a predetermined number of input neurons. Therefore, as a preprocessing of the data, all images are scaled to one resolution before being sent to the network: 512x512 px.
\subsection{Image preprocessing}

When using deep neural networks in image classification problems, additional research is required in order to select the optimal parameters for various parts of the algorithms.
Pre-processing the input data can have a significant effect on training models. Within the framework of this part of the study, the following options for image processing were considered:
\begin{itemize}
    \item Scaling - dividing all values in the image by 255.
    \item Centering the mean of the image at 0 and normalizing the standard deviation at 1.
\end{itemize}
The study tested the following neural network models with standard parameters:
\begin{itemize}
    \item Inception V3, input layer dimension: 299x299
    \item ResNet-50, input layer dimension: 224x224
    \item DenseNet-201, input layer dimension: 224x224
\end{itemize}
All models were trained over 10 epochs, the size of one package was 16 images. Categorical cross entropy was used as the loss function, and Adam was used as the optimizer.
 Learning was stopped in advance if the value of the loss function on the validation set did not decrease for five epochs.
The values of the error function (loss), precision and recall metrics based on training and validation results (prefix ``val\_'') for processed images using scaling and centering methods are indicated in the tables \ref{rescale} and \ref{samplewise}, respectively.

\begin{table}[h]
\centering
\caption{Results of training models with standard parameters and image prescaling.\\ Values in brackets are 95\% confidence intervals.}\label{rescale}
\begin{tabular}{|c|c|c|c|}
\hline
\textbf{Metrics}   & \textbf{InceptionV3}  & \textbf{ResNet-50 V2} & \textbf{DenseNet-201} \\ \hline
\textbf{loss}      & 0.24 {[}0.22, 0.25{]} & 0.32 {[}0.30, 0.35{]} & 0.27 {[}0.25, 0.29{]} \\ \hline
\textbf{precision} & 0.92 {[}0.91, 0.92{]} & 0.89 {[}0.87, 0.90{]} & 0.91 {[}0.89, 0.92{]} \\ \hline
\textbf{recall}    & 0.91 {[}0.91, 0.92{]} & 0.87 {[}0.86, 0.89{]} & 0.90 {[}0.88, 0.91{]} \\ \hline
\textbf{val\_loss} & 0.27 {[}0.27, 0.29{]} & 0.24 {[}0.22, 0.26{]} & 0.32 {[}0.30, 0.34{]} \\ \hline
\textbf{val\_precision} & 0.79 {[}0.79, 0.81{]} & 0.74 {[}0.72, 0.76{]} & 0.76 {[}0.74, 0.78{]} \\ \hline
\textbf{val\_recall}    & 0.77 {[}0.77, 0.79{]} & 0.71 {[}0.69, 0.73{]} & 0.73 {[}0.71, 0.75{]} \\ \hline
\end{tabular}
\end{table}

\begin{table}[h]
\centering
\caption{Results of training models with standard parameters and preliminary centering of images.\\ Values in brackets are 95\% confidence intervals.}\label{samplewise}
\begin{tabular}{|c|c|c|c|}
\hline
\textbf{Metrics}   & \textbf{InceptionV3}  & \textbf{ResNet-50 V2} & \textbf{DenseNet-201} \\ \hline
\textbf{loss}      & 0.34 {[}0.32, 0.36{]} & 0.34 {[}0.31, 0.36{]} & 0.37 {[}0.34, 0.39{]} \\ \hline
\textbf{precision} & 0.89 {[}0.87, 0.90{]} & 0.88 {[}0.87, 0.90{]} & 0.87 {[}0.86, 0.89{]} \\ \hline
\textbf{recall}    & 0.87 {[}0.85, 0.88{]} & 0.87 {[}0.86, 0.89{]} & 0.86 {[}0.85, 0.88{]} \\ \hline
\textbf{val\_loss} & 0.53 {[}0.50, 0.55{]} & 0.93 {[}0.92, 0.94{]} & 0.35 {[}0.33, 0.38{]} \\ \hline
\textbf{val\_precision} & 0.72 {[}0.70, 0.74{]} & 0.65 {[}0.63, 0.67{]} & 0.76 {[}0.75, 0.78{]} \\ \hline
\textbf{val\_recall}    & 0.67 {[}0.65, 0.69{]} & 0.62 {[}0.59, 0.64{]} & 0.71 {[}0.71, 0.75{]} \\ \hline
\end{tabular}
\end{table}




As it can be seen from the tables, pre-scaling of values gives metric values better than centering during training and validation. At the same time, the Inception V3 network showed the highest results in this test.

\subsection{Optimization}
The selected algorithm for optimizing the error function can significantly affect the performance of neural networks. This part of the study analyzes various optimizers: Stochastic Gradient Descent (SGD), RMSProp, and Adam.  Testing was carried out on InceptionV3, ResNet-50, DenseNet-201 networks and pre-scaled input images of 500x500 pixels.

All models were trained for 10 epochs, the size of one package was 8 images. The tables \ref{adam}, \ref{rms} and \ref{sgd} indicate the values of the Precision, Recall and F1 metrics obtained as a result of validation for each class.
\begin{table}[h]
\centering
\caption{Results of training networks with Adam optimizer. \\Values in brackets are 95\% confidence intervals.}\label{adam}
\begin{tabular}{|c|c|c|c|c|}
\hline
\textbf{Models}                        & \textbf{Metrics}   & \textbf{COVID-19} & \textbf{Normal}  & \textbf{Pneumonia} \\ \hline
\multicolumn{1}{|c|}{\multirow{3}{*}{\textbf{InceptionV3}}} & \textbf{precision} & 0.53 [0.51, 0.55] & 0.51 [0.49, 0.53] & 0.59 [0.57, 0.59] \\ \cline{2-5}
\multicolumn{1}{|c|}{}                 & \textbf{recall}    & 0.38 [0.37, 0.40]  & 0.58 [0.57, 0.60] & 0.68 [0.68, 0.70]   \\ \cline{2-5}
\multicolumn{1}{|c|}{}                 & \textbf{f1-score}   & 0.44 [0.42, 0.46]  & 0.54 [0.52, 0.56] & 0.63 [0.62, 0.65]   \\ \hline
\multirow{3}{*}{\textbf{ResNet-50 V2}} & \textbf{precision} & 0.55 [0.55, 0.57]  & 0.51 [0.49, 0.53] & 0.52 [0.52, 0.54]   \\ \cline{2-5}
                                       & \textbf{recall}    & 0.40 [0.40, 0.42]  & 0.53 [0.51, 0.55] & 0.63 [0.61, 0.65]   \\ \cline{2-5}
                                       & \textbf{f1-score}   & 0.46 [0.44, 0.46]  & 0.52 [0.50, 0.54] & 0.57 [0.55, 0.59]   \\ \hline
\multirow{3}{*}{\textbf{DenseNet-201}} & \textbf{precision} & 0.52 [0.50, 0.54]  & 0.53 [0.51, 0.55] & 0.53 [0.51, 0.55]   \\ \cline{2-5}
                                       & \textbf{recall}    & 0.56 [0.54, 0.58]  & 0.51 [0.49, 0.53] & 0.51 [0.49, 0.53]   \\ \cline{2-5}
                                       & \textbf{f1-score}   & 0.54 [0.52, 0.56]  & 0.52 [0.50, 0.54] & 0.52 [0.50, 0.54]   \\ \hline
\end{tabular}%
\end{table}

\begin{table}[h]
\centering
\caption{Results of training networks with the RMSprop optimizer.\\ Values in brackets are 95\% confidence intervals.}\label{rms}
\begin{tabular}{|c|c|c|c|c|}
\hline
\textbf{Models}                        & \textbf{Metrics}   & \textbf{COVID-19}  & \textbf{Normal}   & \textbf{Pneumonia} \\ \hline
\multicolumn{1}{|c|}{\multirow{3}{*}{\textbf{InceptionV3}}} & \textbf{precision} & 0.57 [0.55, 0.59] & 0.52 [0.50, 0.54] & 0.53 [0.51, 0.55] \\ \cline{2-5}
\multicolumn{1}{|c|}{}                 & \textbf{recall}    & 0.51 [0.49, 0.53]  & 0.70 [0.68, 0.72] & 0.58 [0.57, 0.60]  \\ \cline{2-5}
\multicolumn{1}{|c|}{}                 & \textbf{f1-score}   & 0.54 [0.52, 0.56]  & 0.60 [0.59, 0.62] & 0.55 [0.55, 0.57]  \\ \hline
\multirow{3}{*}{\textbf{ResNet-50 V2}} & \textbf{precision} & 0.56 [0.54, 0.58]  & 0.63 [0.61, 0.65] & 0.54 [0.52, 0.56]  \\ \cline{2-5}
                                       & \textbf{recall}    & 0.84 [0.84, 0.85]  & 0.36 [0.35, 0.38] & 0.49 [0.48, 0.51]  \\ \cline{2-5}
                                       & \textbf{f1-score}   & 0.67 [0.66, 0.69]  & 0.46 [0.43, 0.48] & 0.51 [0.49, 0.53]  \\ \hline
\multirow{3}{*}{\textbf{DenseNet-201}} & \textbf{precision} & 0.53 [0.51, 0.55]  & 0.56 [0.54, 0.58] & 0.55 [0.55, 0.57]  \\ \cline{2-5}
                                       & \textbf{recall}    & 0.23 [0.21, 0.24]  & 0.65 [0.62, 0.67] & 0.77 [0.75, 0.79]  \\ \cline{2-5}
                                       & \textbf{f1-score}   & 0.32 [0.31, 0.34]  & 0.60 [0.59, 0.62] & 0.64 [0.61, 0.66]  \\ \hline
\end{tabular}%
\end{table}

\begin{table}[h]
\centering
\caption{Results of training networks with the SGD optimizer.\\ Values in brackets are 95\% confidence intervals.}\label{sgd}
\begin{tabular}{|c|c|c|c|c|}
\hline
\textbf{Models}                        & \textbf{Metrics}   & \textbf{COVID-19} & \textbf{Normal}   & \textbf{Pneumonia} \\ \hline
\multicolumn{1}{|c|}{\multirow{3}{*}{\textbf{InceptionV3}}} & \textbf{precision} & 0.63 [0.61, 0.65] & 0.60 [0.59, 0.62] & 0.67 [0.66, 0.69] \\ \cline{2-5}
\multicolumn{1}{|c|}{}                 & \textbf{recall}    & 0.50 [0.47, 0.52] & 0.64 [0.61, 0.66] & 0.64 [0.61, 0.66]  \\ \cline{2-5}
\multicolumn{1}{|c|}{}                 & \textbf{f1-score}  & 0.56 [0.54, 0.58] & 0.62 [0.69, 0.64] & 0.65 [0.62, 0.67]  \\ \hline
\multirow{3}{*}{\textbf{ResNet-50 V2}} & \textbf{precision} & 0.50 [0.47, 0.52] & 0.56 [0.54, 0.58] & 0.52 [0.50, 0.54]  \\ \cline{2-5}
                                       & \textbf{recall}    & 0.58 [0.57, 0.60] & 0.55 [0.55, 0.57] & 0.45 [0.42, 0.47]  \\ \cline{2-5}
                                       & \textbf{f1-score}  & 0.54 [0.52, 0.56] & 0.55 [0.55, 0.57] & 0.48 [0.45, 0.50]  \\ \hline
\multirow{3}{*}{\textbf{DenseNet-201}} & \textbf{precision} & 0.33 [0.30, 0.35] & 0.00 [0.00, 0.00] & 0.33 [0.30, 0.35]  \\ \cline{2-5}
                                       & \textbf{recall}    & 0.88 [0.86, 0.89] & 0.00 [0.00, 0.00] & 0.11 [0.09, 0.12]  \\ \cline{2-5}
                                       & \textbf{f1-score}  & 0.48 [0.45, 0.50] & 0.00 [0.00, 0.00] & 0.17 [0.15, 0.18]  \\ \hline
\end{tabular}
\end{table}

\section{Results and Discussion}
Tables \ref{adam}, \ref{rms} and \ref{sgd} show that different optimizers are differently effective for different networks, so the ResNet-50 V2 network together with the RMSProp optimization algorithm for the Recall metric was able to identify a larger number of images with COVID-19, while Inception V3 with SGD method, shows the highest results on average across classes, according to the F1 metric.

Thus, in the course of the study, it was revealed that to solve the problem of diagnosing COVID-19 using X-ray images, it is preferable to use the Inception network in conjunction with the SGD optimization algorithm. In this case, the values in the input images are recommended to be reduced to the range [0; 1] by means of scaling.

 In the course of the experiments, models were upgraded for the problem, trained and tested based on the following architectures: Inception, ResNet and DenseNet.
 The accuracy of networks according to the accuracy metric during training and testing is shown in Fig. \ref{train-results}.

 \addtwoimghere{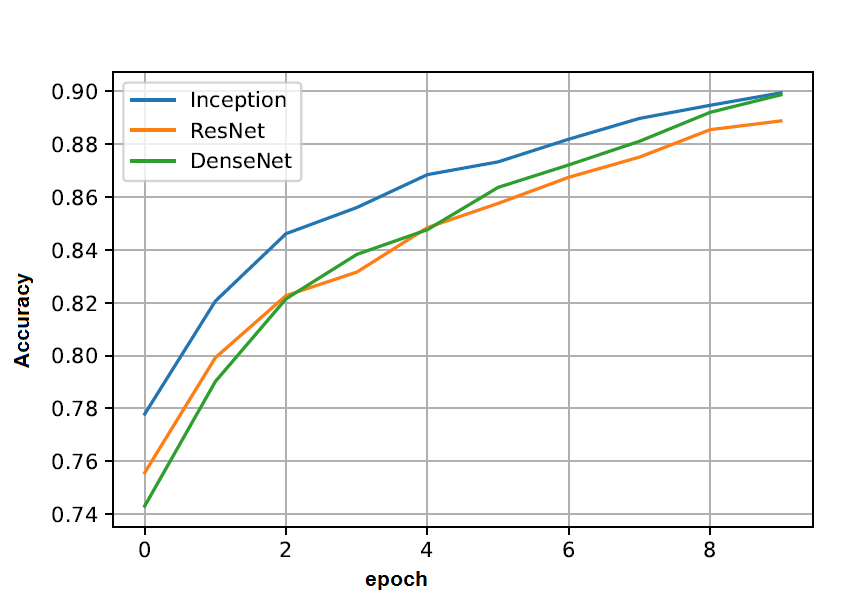}{Training}{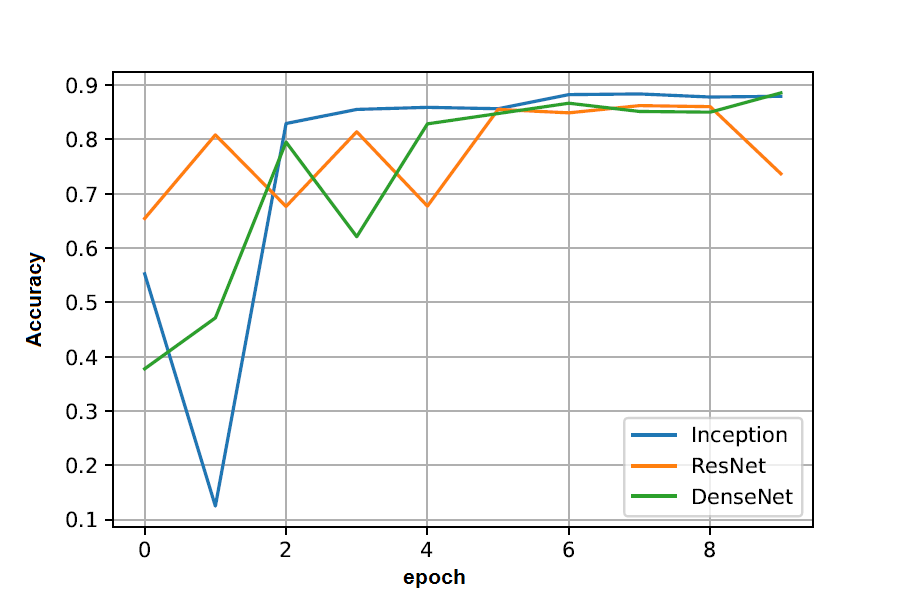}{Testing}{Accuracy of networks}{train-results}

 As seen from Fig. \ref{train-results} the accuracy of the networks increases with the number of training epochs and tends to 1, which indicates the high quality of these algorithms in solving image classification problems.

\section{Conclusion}\label{con}
In this paper, a study of deep machine learning algorithms in image recognition problems was carried out. Convolutional neural network architectures such as Inception, ResNet, and DenseNet were considered. Models were also trained and tested to solve the problem of diagnosing pneumonia and COVID-19 from chest x-rays. The results of the study showed that the Inception V3 network together with the SGD optimization algorithm is better at solving this problem. Preliminary reduction of values to the range [0; 1] can improve the quality of network training. Expanding training data and more epochs can significantly improve the quality of image recognition.

\section*{Compliance with Ethical Standards}
\subsection*{Funding}
 No funding was received for this study.
\subsection*{Conflict of interest}
The authors declare no conflict of interest, financial or otherwise.
\subsection*{Research involving human participants and/or animals}
This research paper does not contain any studies with human participants or animals performed by any of the authors.
\subsection*{Ethical approval}
Not needed.
\subsection*{Informed consent}
All authors agreed on the submitted version.


\bibliographystyle{unsrt}





\end{document}